\def\year{2020}\relax
\documentclass[letterpaper]{article} 
\usepackage{aaai20}  
\usepackage{times}  
\usepackage{helvet} 
\usepackage{courier}  
\usepackage[hyphens]{url}  
\usepackage{graphicx} 
\usepackage{graphicx}  

\usepackage[utf8]{inputenc}

\usepackage[]{todonotes}

\usepackage{amssymb}
\usepackage{amsmath}
\usepackage{amsthm}

\usepackage{booktabs}

\newtheorem{theorem}{Theorem}

\theoremstyle{definition}
\newtheorem{definition}{Definition}

\usepackage[nolist,nohyperlinks]{acronym}
\usepackage[inline]{enumitem}

\usepackage{pgf}
\usepackage{tikz}
\usetikzlibrary{arrows,automata}
\usetikzlibrary{shapes,snakes}
\usetikzlibrary{intersections}

\usepackage{mathtools,algorithm}
\usepackage[noend]{algpseudocode}
\usepackage[capitalise]{cleveref}

\theoremstyle{proposition}
\newtheorem{proposition}{Proposition}

\theoremstyle{corollary}

\crefname{algorithm}{Alg.}{Algs.}
\Crefname{algorithm}{Algorithm}{Algorithms}

\tikzstyle{line} = [draw, -latex']
\tikzstyle{hidden} = [ellipse, draw, text centered, inner sep=1pt]
\tikzstyle{obs} = [ellipse, draw, fill=gray!60, text centered, inner sep=1pt]
\tikzstyle{rv} = [draw, ellipse, inner sep=2pt]
\tikzstyle{pf} = [draw, rectangle, fill=gray]
\tikzstyle{pc} = [draw, rounded corners=15pt, align=center, minimum width=18mm, font=\normalsize, inner sep=3pt]
\tikzstyle{pc2} = [draw, rounded corners=8pt,align=center,minimum height=6mm,font=\normalsize,inner sep=2pt]

\tikzstyle{nhidden} = [draw=none,fill=none,ellipse, text centered, inner sep=1pt]
\tikzstyle{nobs} = [draw=none,fill=none,ellipse, fill=gray!60, text centered, inner sep=1pt]
\tikzstyle{nrv} = [draw=none,fill=none,ellipse, inner sep=2pt]
\tikzstyle{npf} = [draw=none,fill=none,rectangle]

\tikzstyle{ID} = [draw, circle, font=\normalsize]
\tikzstyle{INN} = [draw, circle, inner sep=1pt, fill=black]

\title{Exploring Unknown Universes in Probabilistic Relational Models\thanks{Paper accepted at AI-19 (Braun and M\"oller 2019)}}

\author{Tanya Braun, Ralf M\"oller\\
 Institute of Information Systems, University of L\"ubeck\\
 \{gehrke, moeller\}@ifis.uni-luebeck.de}

\begin{document}

\maketitle

\begin{abstract}
Large probabilistic models are often shaped by a pool of known individuals (a universe) and relations between them.
Lifted inference algorithms handle sets of known individuals for tractable inference. 
Universes may not always be known, though, or may only described by assumptions such as ``small universes are more likely''.
Without a universe, inference is no longer possible for lifted algorithms, losing their advantage of tractable inference.
The aim of this paper is to define a semantics for models with unknown universes decoupled from a specific constraint language to enable lifted and thereby, tractable inference.
\end{abstract}

\section{Introduction}
\nocite{BraMo19a}

At the heart of many machine learning algorithms lie large probabilistic models that use random variables (randvars) to describe behaviour or structure hidden in data.
After a surge in effective machine learning algorithms, efficient algorithms for inference come into focus to make use of the models learned or to optimise machine learning algorithms further \cite{LeC18}.
Often, a model is shaped by a pool of known individuals (constants), i.e., a known universe, and relations between them.
Handling sets of individuals enables tractable inference \cite{NieBr14}. 

Lifting efficiently handles sets of individuals by working with representatives of individuals behaving identically and only looking at specific individuals if necessary.
If modelling, e.g., a possible epidemic depending on how many people are sick, all people being sick behave identically towards an epidemic.
In parametric factors (parfactors), randvars parameterised with logical variables (logvars) compactly represent sets of randvars \cite{Poo03}.
Instead of specifying a factor for each person about how the person being sick affects an epidemic, one parfactor works as a template for all people.
Markov logic networks use first-order logic formulas for compact encoding \cite{RicDo06}.
A known universe means that logvars in parfactors or Markov logic networks have a domain and possibly a constraint restricting domains to certain constants for specific parfactors or formulas.
Lifted inference algorithms such as
\begin{enumerate*} 
	\item lifted variable elimination (LVE) \cite{Poo03,TagFiDaBl13}, 
	\item the lifted junction tree algorithm \cite{BraMo17a}, 
	\item first-order knowledge compilation \cite{BroTaMeDaRa11}, 
	\item probabilistic theorem proving \cite{GogDo11}, or 
	\item lifted belief propagation \cite{AhmKeMlNa13},
\end{enumerate*}
use domains or constraints to determine the number of individuals represented to be able to perform efficient inference.

The question is what to do if the universe is unknown, which makes logvar domains unspecified and constraints empty or not applicable.
In the example about an epidemic, the people who are possibly sick are not known.
The question is not entirely new and an interesting one for diverse research areas:
\mbox{Ceylan et al.} define a semantics for open-world probabilistic databases, keeping a fixed upper bound on domains \cite{CeyDaBr16}.
\mbox{Srivastava et al.} specify first-order open-universe partially observable Markov decision processes to generate strategies based on sampling \cite{SriRuRuCh14}.
\mbox{Milch et al.} study unknown domains in Bayesian Logic, using sampling for approximate inference \cite{MilMaRuSoOnKo05}.
But, the effects of \emph{unknown} finite universes on lifted inference and how to treat unknown universes in lifting have not been discussed.

Therefore, this paper explores lifted inference given models with unknown universes by defining semantics decoupled from a specific constraint language to again enable tractable inference with lifted algorithms.
Decoupling the semantics from the constraint language allows for exploring unknown universes unrestricted by the expressiveness of a specific constraint language.
The semantics is based on constraints over constraints and a set of possible domains, resulting in a variety of interesting new queries that allow for exploring unknown universes as well as checking assumptions about models. 
Additionally, we discuss specifying a distribution over domains, similar to~\cite{MilMaRuSoOnKo05}.
Although the idea behind our approach applies to any formalism and lifted algorithm, we consider parfactors together with LVE since LVE has also been decoupled from the constraint language \cite{TagFiDaBl13}.

The remainder of this paper starts with providing notations and recapping LVE.
Then, we discuss constraints and domains from a generative viewpoint and define semantics.
Finally, we look at query answering for such models.

\section{Preliminaries}
This section specifies notations and recaps LVE.
A running example models the interplay of an epidemic and people being sick, travelling, and being treated. 
Travels spread a disease, making an epidemic more likely.
Treatments combat a disease, making an epidemic less likely.
The example shows a scenario where one is interested in transferring a model to varying domains.

\subsection{Parameterised Models}
Parameterised models are the enclosing formalism for parfactors.
A parfactor describes a function, mapping argument values to real values (potentials).
Parameterised randvars (PRVs) constitute arguments, compactly encoding patterns, i.e., the function is identical for all groundings.
Definitions are based on \cite{TagFiDaBl13}.

\begin{definition}\label{def:univ}
	Let $\mathbf{R}$ be a set of randvar names, $\mathbf{L}$ a set of logvar names, $\Phi$ a set of factor names, and $\mathbf{D}$ a set of constants (universe).
	All sets are finite.
	Each logvar $L$ has a domain $\mathcal{D}(L) \subseteq \mathbf{D}$. 
	A \emph{constraint} is a tuple $(\mathcal{X}, C_{\mathbf{X}})$ of a sequence of logvars $\mathcal{X} = (X_1, \dots, X_n)$ and a set $C_{\mathcal{X}} \subseteq \times_{i = 1}^n\mathcal{D}(X_i)$.
	The symbol $\top$ for $C$ marks that no restrictions apply, i.e., $C_{\mathcal{X}} = \times_{i = 1}^n\mathcal{D}(X_i)$.

	A \emph{PRV} $R(L_1, \dots, L_n), n \geq 0$ consists of a randvar $R \in \mathbf{R}$ possibly combined with logvars $L_1, \dots, L_n \in \mathbf{L}$. 
	If $n = 0$, the PRV is parameterless and constitutes a propositional randvar.
	The term $\mathcal{R}(A)$ denotes the possible values (range) of a PRV $A$. 
	An \emph{event} $A = a$ denotes the occurrence of PRV $A$ with range value $a \in \mathcal{R}(A)$. 
	We denote a \emph{parfactor} $g$ by $\phi(\mathcal{A})_{| C}$ with $\mathcal{A} = (A_1, \dots, A_n)$ a sequence of PRVs, $\phi : \times_{i = 1}^n \mathcal{R}(A_i) \mapsto \mathbb{R}^+$ a function with name $\phi \in \Phi$, and $C$ a constraint on the logvars of $\mathcal{A}$. 
	A PRV $A$ or logvar $L$ under constraint $C$ is given by $A_{|C}$ or $L_{|C}$, respectively.
	We may omit $|\top$ in $A_{|\top}$, $L_{|\top}$, or $\phi(\mathcal{A})_{| \top}$.
	A set of parfactors forms a \emph{model} $G := \{g_i\}_{i=1}^n$.
\end{definition}
The term $lv(P)$ refers to the logvars in $P$, which may be a PRV, a constraint, a parfactor, or a model.
The term $gr(P)$ denotes the set of all instances of $P$ w.r.t.\ given constraints.
An instance is an instantiation (grounding) of $P$, substituting the logvars in $P$ with a set of constants from given constraints.
If $P$ is a constraint, $gr(P)$ refers to the second component $C_{\mathbf{X}}$.
The universe is given by $\mathbf{D}$, and the constraints encode which parfactors apply to which constants.

\begin{figure} 
\centering
\begin{tikzpicture}[rv/.style={draw, ellipse, inner sep=1pt},pf/.style={draw, rectangle, fill=gray},label distance=0.5mm]
	\node[rv] 									(se)	{$Epid$}; 
	\node[pf, right of=se, xshift=0mm, label=0:{$g_0$}]	(seH){};
	\draw (seH) -- (se);
	\node[rv, below of=se, node distance=8mm]		(E)	{$Sick(X)$};
	\node[rv, left of=se, xshift=-18mm, yshift=-4mm]	(R)	{$Travel(X)$};
	\node[rv, right of=se, xshift=19mm, yshift=-4mm]	(S)	{$Treat(X, T)$};
	\node[pf, left of=se, xshift=-3mm, yshift=-3mm]	(REa) {};
	\node[pf, left of=se, label=240:{$g_1$},yshift=-4mm, xshift=-2mm]	(RE) {};
	\draw (se) -- (RE);
	\draw (RE) -- (R);
	\draw (RE) -- (E);
	\node[pf, left of=se, xshift=-1mm, yshift=-5mm]		(REb) {};
	\node[pf, right of=se, xshift=1mm, yshift=-3mm]		(ESa) {};
	\node[pf, right of=se, label=330:{$g_2$},yshift=-4mm, xshift=2mm]	(ES) {};
	\draw (se) -- (ES);
	\draw (ES) -- (E);
	\draw (ES) -- (S);
	\node[pf, right of=se, xshift=3mm, yshift=-5mm]	(ESb) {};
\end{tikzpicture}
\caption{Parfactor graph for $G_{ex}$}
\label{fig:exFG}
\end{figure}
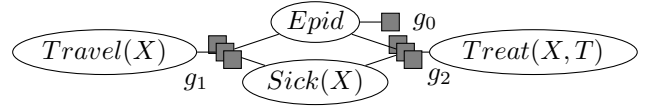

Let us specify a model $G_{ex}$ for the epidemic example.
The sets of names are $\mathbf{R} = \{Epid,$ $Sick, Travel, Treat\}$, $\mathbf{L} = \{X, T\}$, and $\Phi = \{\phi_0, \phi_1, \phi_2\}$.
The set of constants $\mathbf{D}$ contains constants $alice, bob, eve$ and $serum_1, serum_2$, which form the domains $\mathcal{D}(X) = \{alice, bob, eve\}$ and $\mathcal{D}(T) = \{serum_1, serum_2\}$.
We build the boolean PRVs $Epid, Sick(X), Travel(X), Treat(X, T)$ from $\mathbf{R}$ and $\mathbf{L}$.
$Epid$ holds if an epidemic occurs.
$Sick(X)$ holds if a person $X$ is sick, $Travel(X)$ holds if $X$ travels, and $Treat(X, T)$ holds if $X$ is treated with $T$.
With a constraint $C = (X, \{eve, bob\})$, $gr(Sick(X)_{| C}) = \{Sick(eve), Sick(bob)\}$.
With a $\top$ constraint, $gr(Sick(X)_{| \top})$ contains $Sick(alice)$ as well.
The model is given by $G_{ex} = \{g_i\}_{i=0}^2$, 
\begin{align}
	g_0 =	& \phi_0(Epid),								\label{eq:g0}\\
	g_1 =	& \phi_1(Epid, Sick(X), Travel(X))_{| C_1},			\label{eq:g1} \\
			&C_1 = \top = \mathcal{D}(X) ,					\nonumber\\  
	g_2 =	&\phi_2(Epid, Sick(X), Treat(X,T))_{| C_2},		\label{eq:g2} \\
			&C_2 = \top = \mathcal{D}(X)\times \mathcal{D}(T).	\nonumber
\end{align}
Parfactors $g_1$ and $g_2$ have eight input-output pairs, $g_0$ has two (omitted here). 
Constraints are $\top$, meaning, the $\phi$'s apply to all possible groundings of the argument PRVs, e.g., $gr(g_1)$ contains three factors, one for $alice, bob, eve$ each, with identical $\phi_1$.
\Cref{fig:exFG} depicts $G_{ex}$ as a graph with four variable nodes for the PRVs and three factor nodes for the parfactors with edges to arguments.

The \emph{semantics} of a model $G$ is given by grounding and building a full joint distribution $P_G$.
Query answering refers to computing probability distributions, which boils down to computing marginals on $P_G$.
A formal definition follows.
\begin{definition}\label{def:qa}
	With $Z$ as normalising constant, a model $G$ represents the full joint distribution $P_G = \frac{1}{Z} \prod_{f \in gr(G)} f$ \emph{(distribution semantics)}.
	The term $P(\mathbf{Q} | \mathbf{E})$ denotes a \emph{query} in $G$ with $\mathbf{Q}$ a set of grounded PRVs and $\mathbf{E}$ a set of events.
\end{definition}
An example query for $G_{ex}$ is $P(Epid | Sick(eve) = true)$, asking for the conditional distribution of $Epid$ given the event $Sick(eve) = true$. 
Lifted query answering algorithms like LVE seek to avoid grounding and building $P_G$. 

\subsection{Lifted Variable Elimination: An Example}
LVE answers queries of the form in \cref{def:qa} by eliminating all PRVs that do not occur in a query.
We use LVE as a means to illustrate how known universes are required for calculations. 
The exact workings of LVE are not necessary for understanding the contributions of this paper.

When eliminating a PRV, LVE in essence computes variable elimination for a representative and exponentiates the result for indistinguishable instances (lifted summing out).
While the main idea is rather straightforward, a correct implementation is more involved.
See \cite{TagFiDaBl13} for details on LVE for models of \cref{def:univ}.

To illustrate the effects of a universe, consider a query $P(Epid)$ in model $G_{ex}$.
LVE eliminates the PRVs $Treat(X,T)$, $Travel(X)$, and $Sick(X)$.
To eliminate $Treat(X,T)$ from parfactor $g_2 = \phi_2(Epid, Sick(X), Treat(X, T))_{|\top}$, LVE looks at the constraint of $g_2$, which is $\top$, i.e., $\mathcal{D}(X) \times \mathcal{D}(T)$.
Eliminating $Treat(X,T)$ leaves $X$ as the only logvar in $g_2$.
As such, there must exist the same number of $T$ constants given each $X$ constant for lifted summing out to apply.
For each $X$, there exist \underline{two} $T$ constants, i.e., $serum_1$ and $serum_2$.
Thus, LVE is able to eliminate $Treat(X,T)$ by summing out $Treat(X,T)$ from $\phi_2$ using propositional variable elimination, leading to a parfactor $g_2' = \phi_2'(Epid, Sick(X))_{|\top}$, and then taking each potential in $g_2'$ to the power of \underline{$2$}, leading to $g_2''$.
The $\top$ constraint in $g_2''$ only refers to the domain of $X$.
(On the propositional level, two $Treat$ randvars are eliminated from two $\phi_2$ factors for each $X$ constant and then multiplied.)

Next, LVE eliminates $Travel(X)$ from parfactor $g_1$, which leads to a parfactor $g_1' = \phi_1'(Epid, Sick(X))_{|\top}$, where each potential is taken to the power of $1$ as eliminating $Travel(X)$ does not eliminate a logvar (afterwards $X$ is still part of $g_1'$).
For eliminating $Sick(X)$, LVE multiplies $g_1'$ and $g_2''$ into $g_{12} = \phi_{12}(Epid, Sick(X))_{|\top}$, sums out $Sick(X)$ from $g_{12}$ as in propositional variable elimination.
Summing out $Sick(X)$ eliminates $X$ as well, which requires the potentials after summing out to be taken to the power of \underline{$3$} for the \underline{three} constants $alice, bob, eve$ in the domain of $X$.
The result is then a parfactor with $Epid$ as argument, which LVE multiplies with $g_0$.
The result is a parfactor that contains the queried probability distribution after normalisation.

To determine exponents for sum-out operations, constraints based on a universe are necessary.
Other lifted algorithms need a universe similar to LVE.
E.g., first-order knowledge compilation builds a tree-like helper structure for efficient answering of multiple queries, which contains nodes that represent isomorphic subtrees and requires the number of subtrees represented during calculations \cite{BroTaMeDaRa11}.
The lifted junction tree algorithm builds another form of helper structure for efficiently answering multiple queries using LVE as a subroutine \cite{BraMo17a}.

\section{Models with Unknown Universes}
This section focusses on models with unknown universes.
Constraints over constraints describe possible universes, decoupled from a specific constraint language.
Based on domain and constraint descriptions, we define semantics. 

\subsection{Template Models}
Parameterised models contain constraints that restrict logvars in a parfactor to constants from a known universe.
Without a known universe, the set of constants $\mathbf{D}$ becomes empty.
As a consequence, logvar domains are empty as the domains are defined as subsets of $\mathbf{D}$.
In turn, constraints are no longer defined since they are combinations of subsets of domains.
Last, semantics lose its meaning as it involves grounding a model, which is not possible without constraints. 

We assume, though, that the model itself accurately describes relations.
Thus, a parameterised model without $\mathbf{D}$ and empty constraints becomes a template model that specifies local distributions for unknown instances of PRVs.
\begin{definition}
	A \textbf{\emph{template model}} $\mathcal{G}$ is a set of parfactors $\{\tilde{g}_i\}_{i=1}^n$, in which each $\tilde{g}_i = \phi_i(\mathcal{A}_i)_{|C}$ has an \textbf{\emph{empty}} constraint $C = (\mathcal{X}, C_{\mathcal{X}})$ with $C_{\mathcal{X}} = \bot$. 
\end{definition}
Replacing the constraint in $g_1$ with $((X), \bot)$ and in $g_2$ with $((X{,}T), \bot)$ in $G_{ex}$, template model $\mathcal{G}_{ex} {=} \{\tilde{g}_i\}_{i=0}^2$ arises.
$\mathcal{G}_{ex}$ no longer refers to a specific universe, allowing for using varying numbers of people of treatments.

\subsection{Worlds of Constraints}
With an unknown universe, we implicitly specify constraints through a set of rules that generate tuples for constraints given a specific domain at a later point.
Constraints over constraints enables us to describe how universes arise independent of specific constants.
To model constraints, one could use, e.g., answer set programming \cite{BreEiTr11}, probabilistic Datalog \cite{Fuh95}, ProbLog \cite{RaeKiTo07}, or Bayesian Logic \cite{MilMaRuSoOnKo05}, with the latter three leading to probabilities associated with constraints.
\begin{definition}
	Given a template model $\mathcal{G}$ and a domain set $D$ for $lv(\mathcal{G})$, a \textbf{\emph{constraint program}} $\mathcal{C}$ returns a ordered set of constraint sets $\mathbf{C} = \{\{C_{j,i}\}_{i=1}^n\}_{j=1}^m$, i.e., $\mathcal{C}$ generates a constraint for each parfactor in $\mathcal{G}$.
	We call each generated constraint set $\{C_{j,i}\}_{i=1}^n$ a \textbf{\emph{constraint world}} $CW_j$.
	If $\mathcal{C}$ assigns a probability distribution over all $CW_j$, $\mathcal{C}$ returns an ordered set of tuples $\mathbf{C} = \{(\{C_{j,i}\}_{i=1}^n, p_j)\}_{j=1}^m$ of constraint sets and corresponding probabilities, forming a \emph{distribution over constraint worlds}.
	Instantiating $\mathcal{G}$ with $CW_j$, i.e., replacing empty constraints with the constraints in $CW_j$, yields a parameterised model $G_{|CW_j}$. 
\end{definition}
Let us look at possible constraint programs to illustrate how constraint worlds arise.
The shorthand $\top$ already defines a constraint program $\mathcal{C}^\top$ that generates tuples by building Cartesian products given domains. 
$\mathcal{C}^\top$ generates exactly one constraint world.
Given $\mathcal{G}_{ex}$, $\mathcal{C}^\top$ returns $\{\{C_{1},C_2\}\}$ if $\mathcal{D}$ contains the domains $\mathcal{D}(X) = \{alice, bob, eve\}$ and $\mathcal{D}(T) = \{serum_1, serum_2\}$. 
For a more complex example, assume that there are three treatments $t_1$, $t_2$, $t_3$ with only two treatments applicable at a time, i.e., $\mathcal{D}(T) = \{t_1,t_2,t_3\}$ and $\mathcal{D}(X)$ unknown. 
Each combination has a different probability, e.g., $0.7$ for $(t_1, t_2)$, $0.2$ for $(t_2, t_3)$, and $0.1$ for $(t_1, t_3)$.
A probabilistic Datalog program captures this setup as follows:{\medskip}\\
\texttt{
	\hspace*{1mm} element\_of\_C2(X,Y1) :- linked(X,Y1,Y2).			\\
	\hspace*{1mm} element\_of\_C2(X,Y2) :- linked(X,Y1,Y2).			\\
	\hspace*{1mm} linked(X,Y1,Y2) :- instance\_of\_X(X) \\
	\hspace*{41mm} \& pair(Y1,Y2).	\\
	\hspace*{1mm} 0.7 pair(t1,t2).\\
	\hspace*{1mm} 0.2 pair(t2,t3).\\ 
	\hspace*{1mm} 0.1 pair(t1,t3).		
}{\medskip}

The first three lines denote rules according to which one can generate $(X,T)$-tuples.
The last line denotes probabilistic facts that are disjoint, with probabilities adding up to $1$, to model the combination of treatments.
If given a domain such $\{alice, bob, eve\}$ for $X$, one can add corresponding facts to the program:{\medskip}\\
\texttt{
	\hspace*{1mm} instance\_of\_X(alice).\\
	\hspace*{1mm} instance\_of\_X(bob). \\
	\hspace*{1mm} instance\_of\_X(eve).	
}{\medskip}\\
Asking the queries \texttt{?- element\_of\_C2(X,Y)} and \texttt{?- instance\_of\_X(X)} generates tuples for the constraints in $\mathcal{G}_{ex}$.
Using \texttt{0.7 pair(t1, t2)}, the program returns the following facts, which contain tuples for the constraints in $\mathcal{G}_{ex}$:{\medskip}\\
\texttt{
	\hspace*{1mm} instance\_of\_X(alice).\\
	\hspace*{1mm} instance\_of\_X(bob).\\
	\hspace*{1mm} instance\_of\_X(eve).\\
	\hspace*{1mm} 0.7 element\_of\_C2(alice,t1).\\
	\hspace*{1mm} 0.7 element\_of\_C2(alice,t2).\\
	\hspace*{1mm} 0.7 element\_of\_C2(bob,t1).\\ 
	\hspace*{1mm} 0.7 element\_of\_C2(bob,t2).\\
	\hspace*{1mm} 0.7 element\_of\_C2(eve,t1).\\
	\hspace*{1mm} 0.7 element\_of\_C2(eve,t2).		
}{\medskip}\\
The Datalog program as constraint program $\mathcal{C}^{DL}$ returns three constraint worlds $\{(\{C_{j,i}\}_{i=1}^2, p_j\}_{j=1}^3$ with $p_1 = 0.7$, $p_2 = 0.2$, and $p_3 = 0.1$ and constraints
\begin{align*}
	C_{1,1} &= C_{2, 1} = C_{3, 1} = ((X), \{(alice), (bob), (eve)\})	\\
	C_{1,2} &= ((X, T), \{(alice, t1), (alice, t2), (bob, t1),\\&\hspace*{19.3mm} (bob, t2), (eve, t1), (eve, t2)\})	\\
	C_{2,2} &= ((X, T), \{(alice, t2), (alice, t3), (bob, t2),\\&\hspace*{19.3mm} (bob, t3), (eve, t2), (eve, t3)\})	\\
	C_{3,2} &= ((X, T), \{(alice, t1), (alice, t3), (bob, t1),\\&\hspace*{19.3mm} (bob, t3), (eve, t1), (eve, t3)\})
\end{align*}

A set of constraint worlds yields a set of parameterised models, which inherits the distribution over the set of constraint worlds if existing.
\begin{proposition}\label{prop:cw}
	Let a constraint program $\mathcal{C}$ generate a set of constraint worlds $\{(CW_j, p_j)\}_{j=1}^m$.
	Instantiating a template model $\mathcal{G}$ with each constraint world $CW_j \in \{(CW_j, p_j)\}_{j=1}^m$ leads to a distribution over the ordered set of parameterised models $\{(G_{|CW_j}, p_j)\}_{j=1}^m$. 
	If  $\mathcal{C}$ does not generate probabilities, the implicit distribution is a uniform distribution with $\forall j : p_j = \frac{1}{m}$.
\end{proposition}
\Cref{prop:cw} relies on $CW$ being valid for $\mathcal{G}$, meaning, $\mathcal{C}$ generates fitting constraints for all parfactors.
Regarding our example, $\mathcal{C}^{DL}$ generates three constraint worlds, each with two constraints, to instantiate $\mathcal{G}_{ex}$.
Using rules in a constraint program is a form of meta-level logic programming, which allows for formulating constraints on constraints without a specific domain.

Next, we consider possible domains and distributions over domains.

\subsection{Worlds of Domains}
Constraint programs still need domains or constants to generate constraint worlds.
In unknown universes, these constants are not available.
In a naive way, one could generate all possible domains, from one constant for each logvar to infinite domains, leading to infeasibly many possible domains. 
Given knowledge about the setting in which one wants to reason (like in the example above about treatments $t_1, t_2, t_3$), one may list all possible domains.
Assumptions may further limit the number of worlds, e.g.:
\begin{enumerate*}
	\item Logvars require discrete domains of at least one element. 
	\item Small worlds (domains) are usually more likely than large ones. 
	\item Only ``orders'' of domain sizes are relevant, not a set of domain sizes with an increment of $1$ between them.
\end{enumerate*}
Depending on the concrete use case, setting up a discrete distribution over domain sizes might be valuable, with the distribution depending on assumptions valid for the use case.
\begin{definition}
	Given a template model $\mathcal{G}$, a \textbf{\emph{domain world}} $DW$ is a set of domains $\{\mathcal{D}(X)\}_{X \in lv(\mathcal{G})}$ for $\mathcal{G}$.
	Given a set of domain worlds $\{DW_k\}_{k=1}^l$ and probabilities $p_k$ for each $DW_k$ s.t.\ $\forall k : p_k \in [0, 1]$ and $\sum_{k} p_k = 1$, then $\mathbf{D} = \{(DW_k, p_k)\}_{k=1}^l$ forms a \emph{distribution over domain worlds}.
	Providing a constraint program $\mathcal{C}$ with $DW_k$ yields a set of constraint worlds $\{CW_j\}_{j=1}^m$.
	Instantiating $\mathcal{G}$ with $\{CW_j\}_{j=1}^m$ yields a set of parameterised models $\{G_{|DW, CW_j}\}_{j=1}^m$.
\end{definition}

One may start with a set of guaranteed constants and add varying numbers of possible constants for domain worlds, inspired by the $\lambda$-completion of open-world probabilistic databases \cite{CeyDaBr16}.
The probabilities allow for measuring how likely a particular instantiation is compared to others.
Given a distribution, one can specify a threshold $t$ to account only for domains with a probability larger $t$, which enables some filtering even before generating parameterised models for efficiency.
Another way of restricting the number of worlds is to take domains that lie within the standard deviation from the mean or those whose probability make up around $95\%$ of the distribution around its mean or maximum value.

\begin{figure} 
\centering
	\includegraphics[width=.48\textwidth]{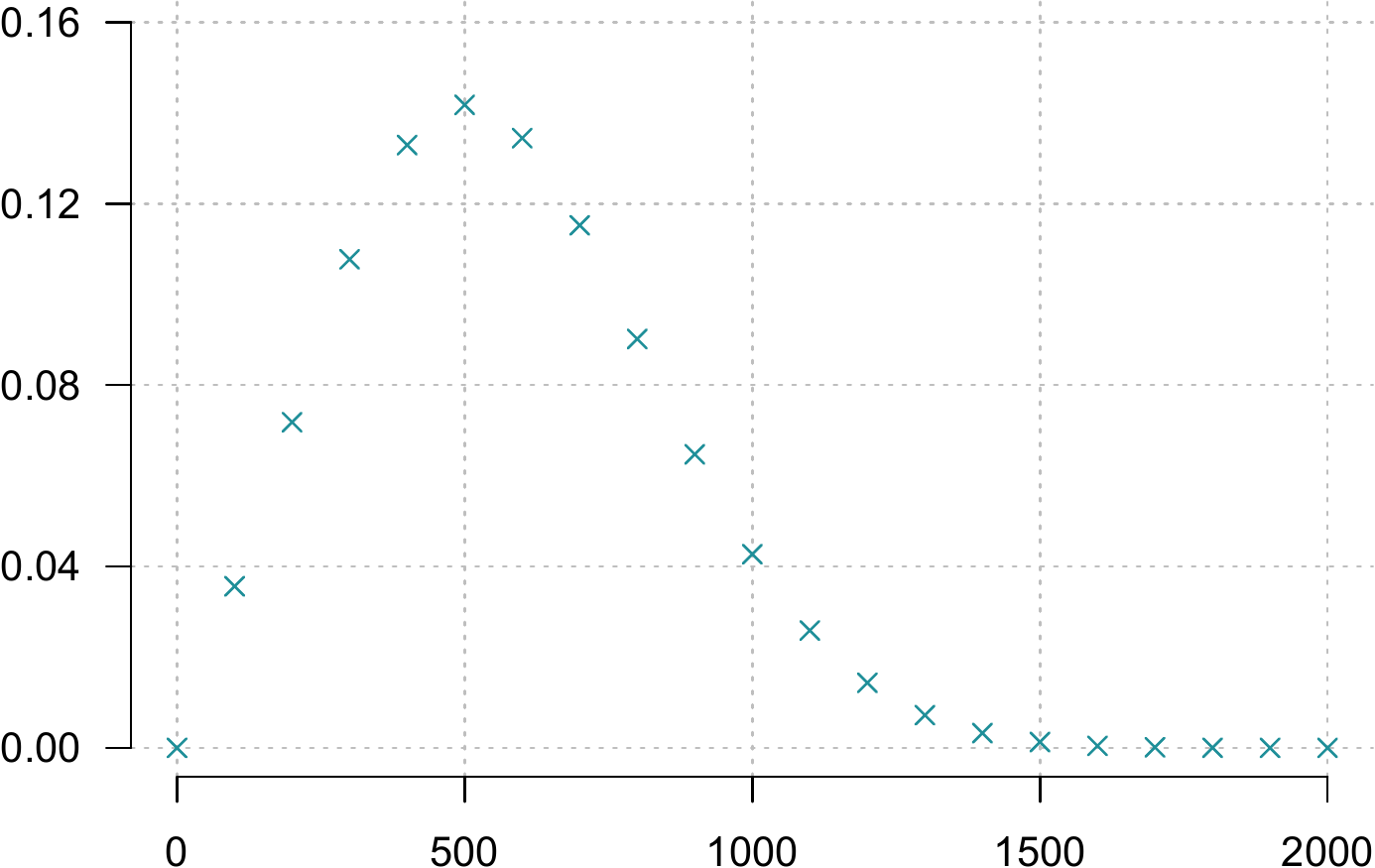} 
	\caption{Discrete distribution over domain sizes of a logvar}
	\label{fig:unkn:distr}
\end{figure}

Let us consider an example distribution for a single logvar, e.g., $X$, the only unknown logvar given $\mathcal{G}_{ex}$ and $\mathcal{C}^{DL}$.
\Cref{fig:unkn:distr} shows a beta-binomial distribution ($\alpha = 6$, $\beta = 15$) based on the assumptions above.
Possible domain sizes $d$ go from $0$ to $2000$ with a step size of $100$ and probabilities for $[d-100, d]$ for $d>0$.
A domain size of $0$ has a probability of $0$.
The highest probability lies with a domain size of $500$, after which probabilities decrease again.
The probability of a domain size of $2000$ is around $3.85 \cdot 10^{-7}$.
Probability distributions between domain and constraint worlds are joined as follows. 
\begin{proposition}\label{prop:dw}
	Let $\{(DW_k, p_k)\}_{k=1}^l$ form a distribution over domain worlds $DW_k$.
	Providing a constraint program $\mathcal{C}$ with $DW_k$ leads to a set of constraint worlds $\mathbf{C}_k = \{(\{C_{k,j,i}\}_{i=1}^n, p_k \cdot p_j)\}_{j=1}^m$ in which $p_j = \frac{1}{m}$ if $\mathcal{C}$ does not assign probabilities. 
	If $\mathcal{C}$ assigns probabilities but only a set of domains $\{DW_k\}_{k=1}^l$ is given, $\{DW_k\}_{k=1}^l$ is extended to form a distribution by setting $\forall k : p_k = \frac{1}{l}$.
\end{proposition}
Multiplying probabilities $p_j$ and $p_k$ relies on $p_j$ and $p_k$ being independent.
The independence assumption is reasonable given the discourse so far as the domain world probability does not influence the generation of constraint worlds, which allows for multiplying the probabilities of domain world and constraint world.
Otherwise, the product has to be replaced with an appropriate expression.
Assigning a probability distribution over possible worlds follows Bayesian thinking, which considers all possible worlds.
Restricting a model to one possible world (with probability $1$) is a simplification, which our approach resolves.

Passing on a domain world to a constraint program $\mathcal{C}$ enables $\mathcal{C}$ to generate constraint worlds for a template model.
Given $\mathcal{G}_{ex}$ and $\mathcal{C}^{DL}$, assume the distribution from \cref{fig:unkn:distr} for $X$, denoted by  $p_x(d)$ with $d$ referring to the domain size of $X$.
There are $20$ domain worlds $\mathbf{D}^{ex} = \{(\{x_i\}_{i=1}^{d}, p_x(d)\}_{d=100, d+=100}^{2000}$ with probabilities $p_x(d)$ between $3.85 \cdot 10^{-7}$ and $1.42 \cdot 10^{-1}$.
For each domain world, $\mathcal{C}^{DL}$ yields three constraint worlds $\{(\{C_{d,j,i}\}_{i=1}^2, p_x(d) \cdot p_{j})\}_{j=1}^3$, i.e., overall $60$ constraint worlds, each containing a constraint for both $\tilde{g}_1$ and $\tilde{g}_2$.
Some of the $60$ constraint worlds have very small probabilities.
Hence, one could use a threshold of $t = 0.05$ to restrict the domain worlds in $\mathbf{D}^{ex}$ to use as inputs for $\mathcal{C}^{DL}$.
Given the distribution of \cref{fig:unkn:distr}, $t$ restricts the domain to sizes between $200$ and $900$, which would lead to $8 \cdot 3 = 24$ constraint worlds.
One could cascade the filtering and drop constraint worlds if their probability goes below $t$ as well (or choose a new $t$).
Given $\mathbf{D}^{ex}$ as an input to $\mathcal{C}^{DL}$ and $t = 0.05$ for cascaded filtering, the number of constraint worlds goes down to $7$, i.e., domain sizes $200$ to $800$ combined with \texttt{0.7 pair(t1,t2)}. 
The constraint worlds using \texttt{0.2 pair(t2,t3)} and \texttt{0.1 pair(t1,t3)} have a probability below $t$.
With domain and constraint worlds in place, we define a semantics for models with unknown universes.

\subsection{Distribution-based Semantics}
To fully specify a model with an unknown universe, we require three components:
\begin{enumerate*}
	\item A \emph{template model} $\mathcal{G}$ provides a structure and local distributions.
	\item A \emph{constraint program} $\mathcal{C}$ generates constraint worlds. 
		A template model can be instantiated with a constraint world, leading to a parameterised model as in \cref{def:univ}, which follows distribution semantics.
	\item A \emph{set of domain worlds} $\mathbf{D}$ specifies (a distribution over) possible domain worlds.
		Each domain world can be passed to the constraint program.
\end{enumerate*}
The semantics are defined as follows.
\begin{definition}
	Let $\mathcal{G}$ be a template model, $\mathcal{C}$ a constraint program, and $\mathbf{D}$ domain worlds.
	A model with unknown universe is given by a triple $(\mathcal{G}, \mathcal{C}, \mathbf{D})$.
	The semantics is given by instantiating $\mathcal{G}$ with constraint worlds $\mathbf{C}$ for each $DW \in \mathbf{D}$.
	The result is a set of parameterised models $\mathbf{G} = \{(G_{|CW}, p)\}_{CW \in \mathcal{C}(DW), DW \in \mathbf{D}}$. 
\end{definition}

Using the formalism of a constraint program, decoupled from a specific constraint language, allows for choosing a constraint language suitable for a specific setup. 
One could use Bayesian logic to specify a distribution over possible models~\cite{MilMaRuSoOnKo05}. 
Using parameterised models as a basis makes it straightforward to retain the capability for lifted inference, especially exact inference. 

The section above discusses the constraint worlds coming from domain worlds, which in turn lead to parameterised models:
With $\mathcal{G}_{ex}$, $\mathcal{C}^{DL}$, $\mathbf{D}^{ex}$, and cascading filtering with $t=0.05$, the semantics yields eight constraint worlds $\mathbf{C}^{ex5} = \{(\{C_{d,j=1,i}\}_{i=1}^2, p_x(d) \cdot p_{j=1})\}_{d=200, d+=100}^{800}$, leading to parameterised models $\mathbf{G}_{ex} = \{(G_{ex|CW_{d,1}}, p_x(d) \cdot p_{1})\}_{d=200, d+=100}^{800}$.
Each $G \in \mathbf{G}_{ex}$ contains parfactors $g_0, g_1, g_2$ with signatures as in \cref{eq:g0,eq:g1,eq:g2} and identical mappings.
Constraints $C_1$ and $C_2$ as well as associated probabilities differ between the models.
For $d=100$, the probability is $3.56 \cdot 10^{-2} \cdot 0.7$ and the constraints are
\begin{align*}
	C_1 &= ((X), \{(x_1), \dots, (x_{100})\}), \\
	C_2 &= ((X, T), \{(x_1, t_1), (x_1, t_2) \dots (x_{100}, t_1), (x_{100}, t_2)\}).
\end{align*}
A domain size of $d=500$ leads to the most probable model.
The last step on our mission of exploring unknown universes is query answering.

\begin{figure*}
\centering
	\includegraphics[width=.48\textwidth]{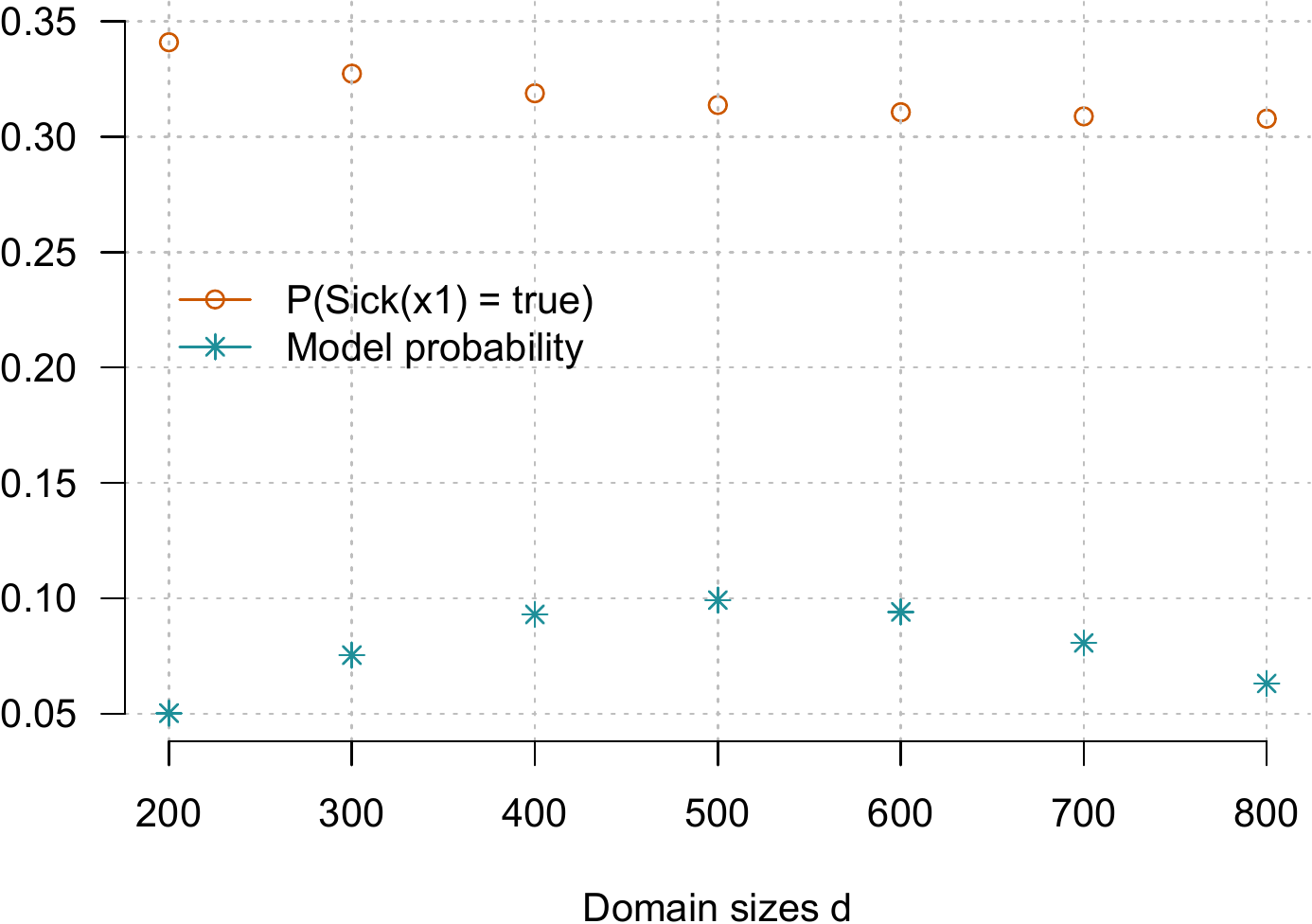}\hfill
	\includegraphics[width=.48\textwidth]{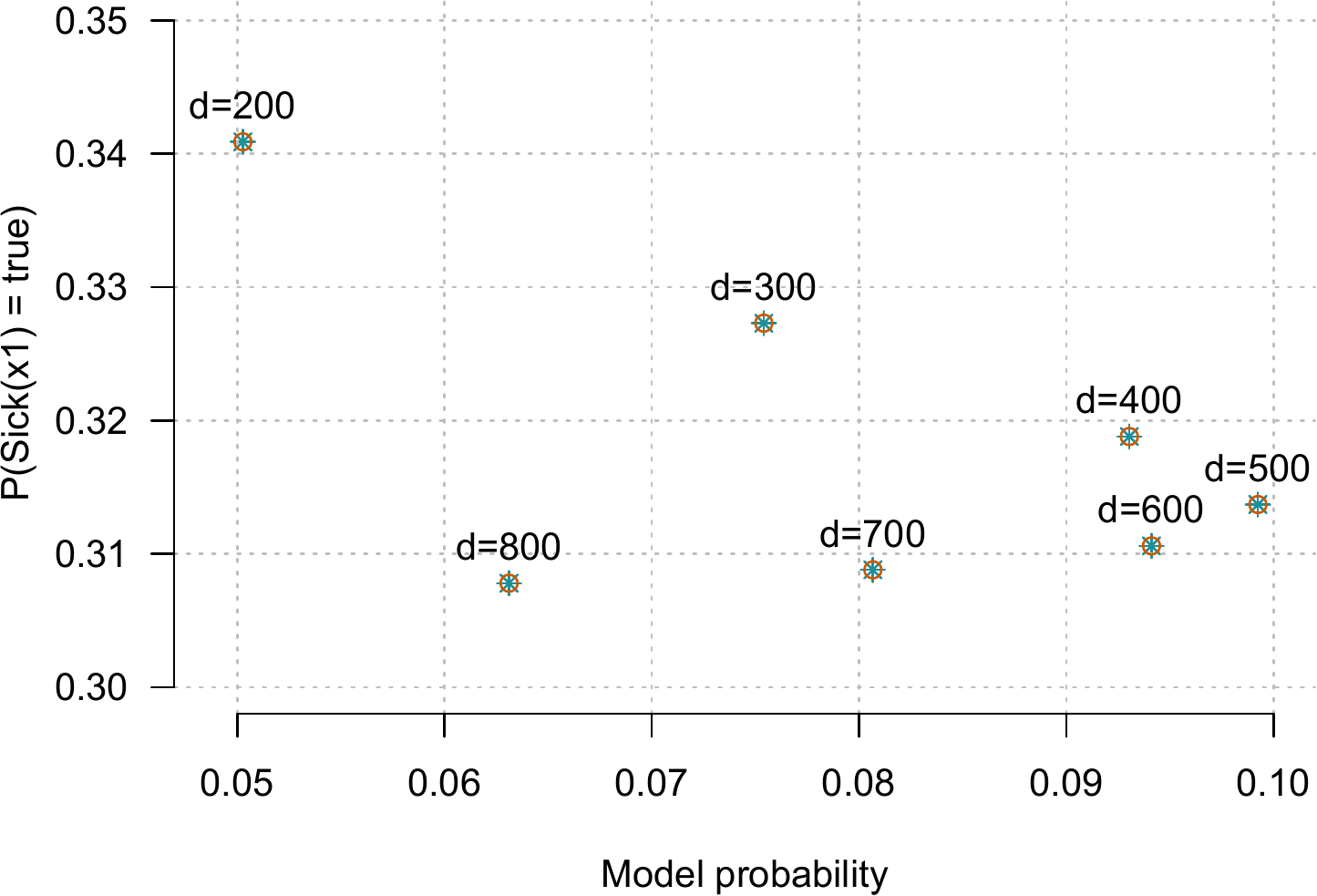}
	\caption{Left: $P(Sick(x_1) = true)$ and model probability for each parameterised model in $\mathbf{G}_{ex}$. Right: Model probability and $P(Sick(x_1) = true)$ plotted for a Skyline query.}
	\label{fig:unkn:qa}
\end{figure*}

\section{Query Answering in Unknown Universes}
The semantics of a model with an unknown universe yields a set of parameterised models.
In each parameterised model, query answering works as before, using LVE (or any other algorithm of one's liking) to answer queries, reaching a main goal of this paper, again enabling tractable inference.
\begin{theorem}\label{thm:lift}
	Given a template model $\mathcal{G}$, a constraint program $\mathcal{C}$ for $\mathcal{G}$, and a set of domain worlds $\mathbf{D}$ for $\mathcal{G}$, resulting in a set of parameterised models $\mathbf{G}$, query answering on each $G \in \mathbf{G}$ is polynomial w.r.t.\ domain-sizes given a domain-lifted inference algorithm, leading to a runtime complexity of $O(|\mathbf{G}| \cdot T_{lift})$ with $T_{lift}$ referring to the runtime complexity of the inference algorithm used.
\end{theorem}
Answering a query on a set of parameterised models $\mathbf{G}$ means that the answer is a set of probabilities or distributions.
If $\mathbf{G}$ has a probability distribution associated, the set of answers has the same distribution associated.
\begin{proposition}\label{prop:qa}
	Answering a query $P(\mathbf{Q}|\mathbf{E})$ on a set of parameterised models $\mathbf{G} = \{G_i\}_{i}$, with $i$ referring to the different models stemming from the domain and constraint worlds, leads to a set of answers $\{P_{G_i}(\mathbf{Q}|\mathbf{E})\}_{i}$.
	If $\mathbf{G}$ has probabilities associated, i.e., $\mathbf{G} = \{(G_i, p_i)\}_{i}$, then the answers have probabilities associated, i.e., $\{(P_{G_i}(\mathbf{Q}|\mathbf{E}), p_i)\}_{i}$, forming a distribution over answers.
\end{proposition}
That is a query leads to a probability distribution over probabilities or probability distributions as a direct consequence of the definitions and \cref{prop:cw,prop:dw}.
Consider a query for a marginal distribution of $Sick(X)$ instantiated with $x_1$.
Each of the parameterised models in $\mathbf{G}_{ex}$ provides an answer, i.e., a marginal distribution for $Sick(x_1)$.
On the left, denoted by a circle, \cref{fig:unkn:qa} shows the probabilities of $Sick(x_1) = true$ for each model with domain sizes on the x-axis. 
The stars denote the probability associated with each parameterised model. 
As mentioned before, the model with domain size $d=500$ is most probable and returns a probability of $0.31$ for $Sick(x_1) = true$. 
Model probabilities decrease to the left and right of $500$.
The queried probability declines with the domain size rising.

\paragraph{Emerging New Queries:}
As we have a set of parameterised models and, therefore, a set of results, new queries emerge.
If asking for the probability of an event, e.g., $Sick(x_1) = true$, one may be interested in those models whose answers have highest probability (top-k query w.r.t.\ query probability).
A top-3 query w.r.t.\ query probabilities in \cref{fig:unkn:qa} returns the models with domain sizes $2$ to $4$ as they lead to the highest probabilities for $Sick(x_1) = true$.
If events such as $Sick(x_1) = true$ have been observed, guaranteed constants are available and a top-k query supports identifying most probable domain sizes for other logvars.
Given the associated probabilities, one may be interested in a top-k query w.r.t.\ model probabilities or in those models that have the highest combined probabilities of event and model (skyline query w.r.t.\ event and model probability).
\Cref{fig:unkn:qa} plots the model probabilities versus the query probabilities.
The skyline consists of the points labeled $d=200$, $d=300$, $d=400$, and $d=500$, which form the outskirt of the points from the origin of the plane.
Asking for distributions, the results over different models might exhibit shifts or clusters worth investigating.
Another new avenue for queries regards checking assumptions about models, e.g., ``Do similar domain sizes lead to similar query results?'' or ``Do query results behave as expected when domain sizes increase (decrease)?''

As shown, given the semantics of models with unknown universe and LVE as the reference algorithm, one can answer various queries. 
Handling unknown universes leads to more work as an algorithm performs query answering for multiple instances, which share certain aspects.
So, while this paper focusses on the semantics, we briefly consider how one would implement it.

\paragraph{Arriving at an Implementation:}
As the model structure is identical for each constraint world and multiple queries probably have to be answered, LVE would perform some calculations multiple times.
One could choose another algorithm to implement the semantics. 
E.g., the lifted junction tree algorithm or first-order knowledge compilation may provide a more suitable setting to answer multiple queries. 
Both algorithms build a helper structure based on the model.
Given that the model structure is the same over different instantiations, helper structures can be reused, constraints adapted as in adaptive inference \cite{AcaIhMeSu08,BraMo18e}, and results of calculations reused to a certain extent \cite{KazPo16a}.
Additionally, one would seek to specify the constraint program in a way that an algorithm can formulate queries about counts for the constraint program, which returns answers ideally without generating extensional constraints.
Given top-k queries w.r.t.\ query probabilities, one would aim at adapting an implementation in the spirit of top-k queries on probabilistic databases as to not evaluate more models than necessary \cite{Fag99}.

\section{Conclusion}
Lifted inference can be restored for models with unknown domains by creating descriptions of possible constraints and domains.
Using those descriptions, one generates worlds to instantiate a template model.
Instantiating a template model yields a set of parameterised models, in which distribution semantics hold again.
With distribution semantics, lifted and thereby, tractable inference w.r.t.\ domains is possible again.
Given a distribution over domain or constraint worlds, the number of worlds can be restricted to a feasible number.
As the same template model is instantiated with different worlds, efficient query answering is possible, reusing helper structures or calculations.
Thus, the proposed semantics seems to be practically useful.
Additionally, new and interesting queries arise that allow for exploring or checking a model.

New inference tasks include automatic generation of instances guaranteed to exist in open universes or learning constraint rules in unknown universes.
Detaching a model from a known universe brings us closer to understanding how transfer learning works:
Transferring a model from one domain to a next opens up possibilities for assumptions changing w.r.t.\ indistinguishable individuals.

\bibliographystyle{aaai}
\bibliography{lifted_inf}

\end{document}